\def\year{2019}\relax
\documentclass[letterpaper]{article} %DO NOT CHANGE THIS
\usepackage{aaai19}  %Required
\usepackage{times}  %Required
\usepackage{helvet}  %Required
\usepackage{courier}  %Required
\usepackage{url}  %Required
\usepackage{graphicx}  %Required
\usepackage{times}
\usepackage{xcolor}
\usepackage{soul}
\usepackage[utf8]{inputenc}
\usepackage[small]{caption}
\usepackage{amsfonts}
\usepackage{amsmath}
\usepackage{bm}
\usepackage{algorithm}% http://ctan.org/pkg/algorithms
\usepackage{algpseudocode}% http://ctan.org/pkg/algorithmicx
\usepackage{lipsum}% http://ctan.org/pkg/lipsum
\usepackage{graphicx}
\usepackage{comment}

\title{Stochastic Variance Reduction for Deep Q-learning}

\author{
Wei-Ye Zhao$^{1}$, 
Xi-Ya Guan$^3$,
Yang Liu$^2$,
Xiaoming Zhao $^2$,
Jian Peng$^2$
\\ 
$^1$ Carnegie Mellon University \\
$^2$ University of Illinois at Urbana-Champaign\\
$^3$ Shanghai Jiao Tong University\\
weiyezha@andrew.cmu.edu,
gxy10080223@sjtu.edu.cn,
liu301@illinois.edu,
xz23@illinois.edu,
jianpeng@illinois.edu
}
\begin{document}

\maketitle

\begin{abstract}
Recent advances in deep reinforcement learning have achieved human-level performance on a variety of real-world applications. However, the current algorithms still suffer from poor gradient estimation with excessive variance, resulting in unstable training and poor sample efficiency. In our paper, we proposed an innovative optimization strategy by utilizing stochastic variance reduced gradient (SVRG) techniques. With extensive experiments on Atari domain, our method outperforms the deep q-learning baselines on $18$ out of $20$ games.
\begin{comment}
Stochastic optimization methods play a crucial role in the success of many machine learning problems. However, it still often suffers from data noise and large variance issue due to random subsampling as well as the challenge of learning rate scheduling, which further leads to unstable and poor convergence rate during training process. We propose a novel optimization algorithm which combines the adaptive learning ability of Adam with SVRG to reduce the variance of gradient direction estimation and further accelerate the convergence rate. We evaluate the performance of our approach on 20 games of the challenging Arcade Learning Environment and report significant improvement in both reward scores and learning speed.
\end{comment}
\end{abstract}

\section{Introduction}
\begin{comment}
Reinforcement learning (RL)~\cite{sutton1998reinforcement} has recently demonstrated its significant performance in finding solutions to solve challenging real-world tasks, such as human-computer interaction~\cite{maes1993learning}, autonomous driving~\cite{dai2005approach}, video games~\cite{mnih2015human}, visual navigation~\cite{zhu2017target}, and goal-oriented autonomous decision making~\cite{frank2006anatomy}. The effective attempt is deep reinforcement learning, combining reinforcement learning and deep learning~\cite{mnih2015human}, where artificial neural networks are used as efficient functional approximators to model complex representations from raw input data and further enable feature extraction and signal perception.
\end{comment}
The recent advances of supervised deep learning methods have tremendously improved the performance on challenging tasks in computer vision, speech recognition and natural language processing. Artificial neural networks is the core idea of deep learning, which is used to model complex hierarchical data abstractions and representations from raw input data. With the help of deep learning, reinforcement learning (RL)~\cite{sutton1998reinforcement} has recently achieved remarkable success on massive real-world applications, such as human-computer interaction~\cite{maes1993learning}, video games~\cite{mnih2015human}, visual navigation~\cite{zhu2017target}, goal-oriented autonomous decision making~\cite{frank2006anatomy} and autonomous driving~\cite{dai2005approach}.

\begin{comment}
The effective attempt is deep reinforcement learning, combining reinforcement learning and deep learning~\cite{mnih2015human}, where artificial neural networks are used as efficient functional approximators to model complex representations from raw input data and further enable feature extraction and signal perception.
\end{comment}

Q-learning~\cite{watkins1992q} is one of the most popular reinforcement learning algorithms, where the policy is learnt by adjusting the parameters at each training iteration to reduce the mean-squared error in the Bellman equation so as to optimize the cumulative future reward, resulting in sequences of well-defined optimization problems. A standard method to solve optimization problems is gradient descent~\cite{kingma2014adam}. Since it is expensive to compute the full expectation in the gradient, stochastic methods are often used to optimize the loss function based on gradients of small batches of samples. Despite these successes, the inaccurate estimation of gradient as well as huge variance arisen from RL training procedure is still the key problem of these stochastic optimization methods, the inexact approximate gradient estimation can be viewed as the distorted gradient direction. In large scale deep Q learning problem, the Q value is represented with deep Q network with proper tuned network parameters. The DQN learning process can be viewed as iteratively optimizing network parameters process according to gradient direction of the loss function at each stage. Therefore, the inexact approximate gradient estimation with a large variance can largely deteriorate the representation performance of deep Q network by driving the network parameter deviated from the optimal setting, causing the large variability of DQN performance. On the other hand, if we assume the network parameter of DQN is $\theta$, the core learning step of deep Q learning is to minimize the gap between the estimated maximum Q value ($y_(s, a)$) given state $s$ and action $a$ and current Q value($Q(s, a; \theta)$) using the operation that $\hat{\theta} = \mathrm{argmin}_\theta \mathbb{E}\lVert y_(s, a) - Q(s, a; \theta)\rVert^2$. It is noteworthy that $\hat{\theta}$ is obtained with gradient descent, thus if the gradient estimation has a large variance, it requires more iterations of $\mathrm{argmin}$ operation such that $\theta$ could reach $\hat{\theta}$, which means large gradient variance will postpone the process when DQN gets local optima.

\begin{comment}
A key problem of these stochastic optimization methods is that the gradient estimates are often with high variance, very noisy and further lead to unstable convergence.
\end{comment}

In this work we address issues that arise from Approximate Gradient Estimation (AGE), and propose Stochastic Variance Reduction for Deep Q-learning (SVR-DQN) optimization to 
%improve
accelerate the convergence for deep Q-learning by reducing the AGE variance. We conduct the AGE variance analysis and theoretically explain how the proposed algorithm addresses them.We evaluate our proposed algorithm using Arcade learning environment~\cite{bellemare2013arcade}. Our experiments show that SVR-DQN optimization algorithm can significantly reduce the delay before the performance gets off the ground, and further lead to aggressive sample efficiency at initial training stage. Our new strategy outperforms Adam in both reward scores and training time on 18 out of 20 games.
% While
\begin{comment}
Although Adam~\cite{kingma2014adam} is a widely used state-of-the-art stochastic optimization algorithm in deep Q-learning achieving robust performance in a broad range of challenging tasks, the parameter update rule of Adam is solely based on gradients of stochastic sample batch, which is inaccurate due to variance caused by random sampling. To better exploit accurate gradient estimation and reduce the negative impact of noisy gradient, we design a new algorithm by leveraging both advantages from stochastic variance reduced gradient descent (SVRG) technique~\cite{johnson2013accelerating} and Adam. 
\end{comment}

\section{Background}
Reinforcement learning (RL) considers agents operating in an uncertain environment, where agents interact with environment to perform sequential actions. At each time step, the agents react according to the observation from environment, receiving a scalar reward from  environment. The RL algorithm aims to search a policy in order to maximize the final cumulative rewards. To be specific, for time step $t = \{1, \dots, T\}$, the agents sample action $a_t \sim \pi(a_t \vert s_t)$ based on observation $s_t$. Then the agents get reward $r_t$ and next step observation $s_{t+1}$ generated. The goal is to maximize $\sum_{t=1}^T \gamma^{t-1} r_t$, where $\gamma$ is the discounting factor for convergence.
% Specifically

\begin{comment}
To be more specific, let us consider the agent operates over time sequence $t \in \{1, 2, ..., T\}$. At each discrete time step $t$, the agent observes a state $s_t$  and 
% choose
chooses an action $a_t \in A$ according to current policy, where the set of possible discrete actions is $A$. The agent will receives a scalar reward and observes a new state $s_{t+1}$ subsequently. The task of reinforcement learning is to find a policy to maximize the expected cumulative reward starting from the initial state.
\end{comment}

\subsection{Q-Learning}
Q-learning is a satisfied method for solving sequential decision problems. It defines optimal value for each action as the expected future rewards when the optimal policy $\pi^\star$ starting from that action, namely

\begin{equation}\label{post-eq111}
Q_{\pi^\star}(s, a) = \mathbb{E}(\sum\limits_{t=1}^T \gamma^{t-1} r_t \vert s_0 = s, a_0 = a, \pi^\star)
\end{equation}

However, it is intractable to compute the optimal $Q_{\pi^\star}$. Instead, we need to estimate optimal action values through temporal difference learning. The parameterized target with $\mathbf{\theta}$ is formalized as following

\begin{equation}\label{post-eq222}
Y_t^Q = r_{t} + \gamma \max_a Q_\pi(s_{t+1}, a, \mathbf{\theta}_t)
\end{equation}

\subsection{Deep Q-Network}
A deep Q network (DQN) is a multi-layered neural network parameterized with $\mathbf{\theta}$ which outputs a vector $Q_\pi(s, \cdot)$ of action values when given a observation state $s$. If observation space has $m$ dimension and action space has $n$ dimension, the DQN is a mapping from $\mathbb{R}^m$ to $\mathbb{R}^n$. The standard method for DQN\cite{mnih2015human} integrates usage of replay buffer and target network. Ror the experience replay, we store observed transition tuples $(s_t, a_t, r_t, s_{t+1})$ and uniformly sample from them in order to break the correlation between tuple pairs, which will enhance the performance of model. Meanwhile, the target network has exactly the same network structure as the online network except for its parameter $\mathbf{\theta}^-$. The parameters $\mathbf{\theta}^-$ will be copied from online network every certain steps so that $\mathbf{\theta} = \mathbf{\theta}^-$, which will be kept unchanged on all other steps, namely

\begin{equation}\label{post-eq333}
Y_t^{DQN} = r_t + \gamma \max_a Q_\pi(s_{t+1}, a, \mathbf{\theta}_t^-)
\end{equation}

On the other hand, the standard Q-learning and DQN, namely Eq.\ref{post-eq222} and Eq.\ref{post-eq333}, maximize $Q$ with the same values both to select and to evaluate an action, making it more likely to select overestimated values. This phenomenon will cause overoptimistic value estimates. In order to resolve the overoptimism, we use Double Q-learning\cite{van2016deep} to get the following modified target:

\begin{equation*}
Y_t^{DoubleQ} = r_t + Q(s_{t+1}, \arg\max_a Q(s_{t+1}, a, \mathbf{\theta}), \mathbf{\theta}^\prime)
\end{equation*}

\begin{comment}
Q-learning is the well established algorithm to solve reinforcement learning tasks, where an action-value function (Q-function) is maintained. Given state $s$, the action-value function provides reward value $Q(s,a)$ for each possible action. The core idea is to estimate optimal future reward using Bellman equation via Q-function. Given policy parameter $\theta_i$ at iteration $i$, the Q-learning updates its policy by optimizing the following loss function via stochastic gradient descent:
\begin{equation}
\sum \limits_{(s_i,a_i,r_i,s_{i+1})\in B}(Q_{\theta}(s_i,a_i)-(r_i + \gamma \max_a Q_{\theta^-}(s_{i+1},a)))^2
\end{equation}
where $B$ denotes a minibatch of 
%state-action-reward-future state-samples,
information tuples, consisting of current state $s_i$ and action $a_i$, reward $r_i$ and next state $s_{i+1}$. $s_{i+1}$ and $r_i$ are obtained after taking action $a_i$, and $\gamma$ is a discount factor. Note $\theta^-$ 
%are
is the parameters used to compute the target value, which 
%are
is updated to $\theta_i$ every certain training steps. 
\end{comment}

\section{Stochastic Variance Reduction for Deep Q-learning Optimization}

Many machine learning problems are considering the a finite-sum optimization problem as following:
\begin{equation}
\min \limits_{w} f(w) = \frac{1}{n} \sum_{i=1}^n f_i(w)
\label{eq2}
\end{equation}
let $w^* = \arg \min_w f(w)$ denote the optimal solution for Eq.\ref{eq2}, a lot of researches in optimization algorithm are motivated to find solution $w$ such that $f(w) - f(w^*) \le \epsilon$. For large-scale problems in form of Eq.\ref{eq2}, randomized variance reduced first-order methods are especially efficient for their low per iteration cost. In order to develop fast stochastic first-order methods, we should make sure that when the 
%iterate
iteration gets closer to optimum, the variance of randomized updating direction decreases.

\subsection{Adam}
An extension to stochastic gradient descent algorithm called Adam~\cite{kingma2014adam} has recently been adopted for a broad range of deep learning models. Adam is an efficient stochastic optimization method to update network weights iteratively based on first-order gradients information.  Once we have gradients of objective function, Adam can adaptively estimate the first and second moments of the gradients, and further compute adaptive learning rate.

\begin{comment}
Suppose $g$ is the gradient of the stochastic objective $f$, and let $1, \dots, t$ be the subsequent time steps, Adam computes the estimates of the 
%$1^{st}$ 
first (the mean) moment $m_t$ and second raw moment (the uncentered variance) $v_t$ of the gradients by:
\begin{equation}
m_t = \beta_1 \cdot m_{t-1} + (1 - \beta_1) \cdot g_t
\end{equation}

\begin{equation}
v_t = \beta_2 \cdot v_{t-1} + (1-\beta_2)\cdot g_t^2
\end{equation}
where $m_t, v_t$ denote first moment and second raw moment respectively, with $\beta_1, \beta_2\in [0,1)$ to control the exponential decay rates of these moving averages. 

Adam derives its careful choice of step size to update parameters from bias-corrected first moment estimate and second raw moment estimate. Let $w_1$, ...,$w_t$ denote the parameters at subsequent time steps, the update rule is defined as:
\begin{equation}
w_t = w_{t-1} - \frac{\alpha \cdot  \frac{m_t}{1-\beta_1^t}}{\sqrt{\frac{v_t}{1-\beta_2^t}} + \epsilon}
\end{equation}
where $\alpha$ is the predefined step size, the terms $\frac{m_t}{1-\beta_1^t}$ and $\frac{v_t}{1-\beta_2^t}$ are the bias-corrected moving averages. Note that Adam's careful choice of step size relates closely with the first-order gradient information $g_t$. The correspondence between $g_t$ and the true gradient direction has a great impact on the performance of Adam.
\end{comment}

Although Adam~\cite{kingma2014adam} is a widely used state-of-the-art stochastic optimization algorithm in deep Q-learning achieving robust performance in a broad range of challenging tasks, the parameter update rule of Adam is solely based on first-order gradient information of stochastic sample batch, which is inaccurate due to variance caused by random sampling. To better exploit accurate gradient estimation and reduce the negative impact of noisy gradient, we design a new algorithm by leveraging both advantages from stochastic variance reduced gradient descent (SVRG) technique~\cite{johnson2013accelerating} and Adam. 

\subsection{Stochastic Variance Reduced Gradient}
Stochastic variance reduced gradient (SVRG)~\cite{johnson2013accelerating} is an explicit variance reduction method for stochastic gradient descent which does not require gradient storage. SVRG enjoys very fast convergence rate using  variate control and can be applied to complex problem such as neural network training. 

To find approximate solution to optimization problem as Eq.\ref{eq2}, a standard method is gradient descent, which is expensive since it requires $n$ 
%derivatives evaluation 
evaluations of derivatives at each iteration. A popular modification is SGD, which reduce the computation cost of standard gradient descent greatly by sub-sampling:
\begin{equation}
w_t = w_{t-1} - \eta_t\cdot\frac{1}{m}\sum_{i=1}^m\nabla f_{i}(w_{t-1},\varphi_t)
\label{eq6}
\end{equation}
where $\varphi_t$ is random variable depending on $w_{t-1}$, $t=1, 2, \dots$ are the subsequence time steps, $m$ is the size of mini-batch sampled from $n$ instances, and the expectation $\mathbb{E}[\sum_{i=1}^m\nabla f_{i_t}(w_{t-1},\varphi_t)|w_{t-1}]=\sum_{i=1}^{m}\nabla f_i(w_{t-1})$. However, large random variances will arise due to the variance of $\sum_{i=1}^m\nabla f_{i_t}(w_{t-1},\varphi_t)$, which slows down the convergence rate.
\begin{comment}
However, great random variances are introduced due the variance of $\sum_{i=1}^m\nabla f_{i_t}(w_{t-1},\varphi_t)$, which slows down the convergence rate. 
\end{comment}

SVRG maintains the snapshot of estimated $\tilde{w}$ which is close to the optimal $w$ every certain iteration. Given the preserved $\tilde{w}$, SVRG pre-calculates an average gradient $\tilde{\mu}=\frac{1}{n}\sum_{i=1}^{B}\nabla f_i(\tilde{w})$ as an anchor point, where $B$ is a subset of training samples. SVRG modifies Eq.\ref{eq6} as:
\begin{equation}
w_t = w_{t-1} - \eta_t \cdot (\frac{1}{m}\sum_{i=1}^{m}\nabla f_i(w_{t-1})-\frac{1}{m}\sum_{i=1}^{m}\nabla f_i(\tilde{w})+\tilde{\mu})
\label{eq7}
\end{equation}

Note that when $w$ is close to $\tilde{w}$, the difference $\nabla f(w) - \nabla f(\tilde{w})$ is small. When both $w_t$ and $\tilde{w}$ converge to the same optimal parameter $w^*$, then $\tilde{\mu} \to 0$ and $\nabla f(w) \to \nabla f(\tilde{w})$. Therefore, the variance of SVRG in update rule Eq.\ref{eq7} is reduced and SVRG can find the more accurate gradient direction estimation.

\subsection{Stochastic Variance Reduced Deep Q-learning Optimization}
In order to improve the performance of Adam optimization, we apply SVRG to find the accurate gradient direction based on the small stochastic training subset and propagate the optimized first-order information to Adam. Our main algorithm, Stochastic Variance Reduction for Deep Q-learning  Optimization (SVR-DQN), is summarized in Algorithm \ref{SVR-DQN}. 

\begin{algorithm*}[h]
  \caption{Stochastic Variance Reduction for Deep Q-learning  Optimization}\label{SVR-DQN}
  \begin{algorithmic}[1]
    \Procedure{Stochastic Variance Reduction for Deep Q-learning  Optimization}{$B,\eta,m,\alpha,\beta_1,\beta_2,b$}
      \State \textbf{Inputs:}
      \State $B$: Training sample batch size
      \State $\eta$: SVRG learning rate
      \State $m$: Number of SVRG inner loop iteration
      \State $\alpha$: Adam stepsize
      \State $\beta_1,\beta_2\in [0,1)$: Exponential decay rate for moment estimates
      \State $b$: mini-batch size
      \State \textbf{Initialization:}
      \State Initialize $\tilde{w}^0 = 0$\quad(Initialize parameter vector)
      \State Initialize $m_0=0$\quad(Initialize first moment vector)
      \State Initialize $v_0=0$\quad(Initialize second moment vector)
      \For{$s=0,1,2,...$}
        \State $B^s=B$ elements sampled without replacement from all training samples\Comment{training sample batch}
        \State Calculate the anchor point: 
        \State $\tilde{\mu}^s=\frac{1}{B}\sum_{i\in B^s}\nabla f_i(\tilde{w}^s)$
        \State $w_0 = \tilde{w}^s$
        \For {$t=1, 2, ..., m$}\Comment{SVRG variance reduction}
          \State Draw a mini-batch $b^t$ uniformly random from $B^s$\Comment{mini-batch}
          \State Reduce variance and update parameter with mini-batch $b_t$:
          \State $w_t = w_{t-1} - \eta(\frac{1}{b}\sum_{i\in b^t}\nabla f_i(w_{t-1})-\frac{1}{b}\sum_{i\in b^t}\nabla f_i(\tilde{w}^s)+\tilde{\mu}^s)$
        \EndFor
      \State Calculate the more accurate estimation of gradient direction from sample batch $B^s$:
      \State $g_{s}=w_m - \tilde{w}^s$
      \State $m_{s+1}=\beta_1 \cdot m_s + (1-\beta_1)\cdot g_s$ (Update biased first moment estimate)\Comment{Adam process}
      \State $v_{s+1}=\beta_2\cdot v_s + (1-\beta_2)\cdot g_s^2$ (Update biased second raw moment estimate)
      \State $\hat{m}_{s+1} = {m_{s+1}}/(1-\beta_1^{s+1})$ (Compute bias-corrected first moment estimate)
      \State $\hat{v}_{s+1} = {v_{s+1}}/(1-\beta_2^{s+1})$ (Compute bias-corrected second raw moment estimate)
      \State $\tilde{w}^{s+1} = \tilde{w}^s - \alpha \cdot {\hat{m}_{s+1}}/(\sqrt{\hat{v}_{s+1}} + \epsilon)$ (Update parameters)
      \EndFor  
      \State \textbf{return} $\tilde{w}^s$
    \EndProcedure
  \end{algorithmic}
\end{algorithm*}
At the beginning of the algorithm, we form a training sample batch $B^s$ sampling from the whole training instances, and fix it for the whole optimization process in $s$-th outer loop. We calculate the average gradient using samples from $B^s$ to perform the current anchor point $\tilde{\mu}^s$, the difference with the standard SVRG process is that we don't use the whole training samples to construct anchor, which is inspired by Sallinen's Practical SVRG~\cite{harikandeh2015stopwasting} that SVRG can solve optimization problem inexactly to calculate $\tilde{\mu}$ with a subset of training examples, and the convergence rate is unchanged. In the inner loop iteration (SVRG variance reduction), we reduce the variance with the average of randomly selected mini-batch $b^t$ from $B^s$ and update parameter according to updating rule Eq.\ref{eq7}, since the usage of individual training sample has a great variance and is also computational inefficient. In order to thoroughly leverage the information from samples in $B^s$ to find optimal gradient direction estimation, the ideal selection of mini-batch size $b$ and SVRG inner loop iteration number $m$ should satisfy the constraint: $b\times m \ge B$.

After SVRG variance reduction process, we have the updated parameter $w_m$ and previous stored snapshot $\tilde{w}^s$, the variance reduce gradient estimation $g_s$ is calculated as $w_m - \tilde{w}^s$. Note that it is unnecessary to rescale $g_s$, since effective step size in Adam is invariant to the scale of the gradients~\cite{kingma2014adam}. With $g_s$ calculated, we follow the standard Adam procedure to construct bias-corrected first moment estimate and second raw moment estimate and further finalize updating parameters for this training iteration leveraging the more accurate gradient direction estimation.

\section{Approximation Gradient Error Variance Reduction}
\subsection{Approximation Gradient Error Variance}

The Approximation Gradient Error(AGE) is the error in the gradient direction estimation of cost function $f(\tilde{w})$, where $\tilde{w}$ is the hyper-parameters of this function, which are optimized with gradient descent methods iteratively by minimize the DQN loss(Algorithm 1 line 29). Given the certain learning samples preserved in experience buffer, the ideal gradient estimation of loss function is supposed to give the accurate learning direction (derivation value) leveraging the current information provided by those learning samples, thus the agent (DQN) can quickly converge to policy optimas by optimizing hyper-parameter at the gradient direction. However, AGE appears in gradient estimation process. 

AGE is a result of several factors: Firstly, the sub-optimality of current hyper-parameters $\tilde{w}$ due to inexact minimization. Secondly, the constrained representing strength of DQN. Thirdly, the limited representation number of the samples we used for deriving the gradients. Lastly, representation error due to unseen(un-stored)  state transitions and policies caused by finite storage of Experience-Replay buffer. The AGE can cause the distortion of the gradient estimation, thus derive the agent policy to a worse one. The AGE can also cause a large variability of DQN performance and postpone the process when DQN gets to local optima. To analyse the AGE variance we first propose the variance of approximation gradient for one single sample. 

We suppose the gradient estimation from one single sample is $\nabla f_i(\tilde{w}) = AGE_i + \nabla f_i(w_*)$, where $i$ is one training sample from the replay buffer, $w_*$ denotes the exact minimized parameters from current stored samples. We also suppose $\nabla f(w_*)$ denotes the optimal gradient direction given current stored samples, $\mathbb{E}(AGE_i) = 0$,   $\mathrm{Var}(AGE_i) = \sigma^2$:
\begin{align*}
 \mathrm{Var}(\nabla f_i(\tilde{w})) &= \mathrm{Var}(AGE_i + \nabla f_i(w_*))\\
 &=\mathrm{Var}(AGE_i) +\mathrm{Var}(\nabla f_i(w_*))\\
 &=\sigma^2.
 \end{align*} 
 
To give the argument of approximation gradient variance reduction, we begin by deriving the bound of the variance of $\nabla f_i(\tilde{w})$, suppose that each $\nabla f_i$ is $L$-Lipschitz continuous:
\begin{align*}
f_i(\tilde{w}) \geq f_i(w_*) + \langle \nabla f_i(\tilde{w}), \tilde{w}-w_*\rangle\\
+ \frac{1}{2L}\lVert \nabla f_i(\tilde{w}) - \nabla f_i(w_*) \rVert ^2 .
\end{align*}
by summing this inequality above over all the training sample $i$, and divide LHS and RHS by $\frac{1}{n}$, we obtain the bound of approximation gradient error variance $\mathrm{Var}(\nabla f_i(\tilde{w}))$ that:
\begin{equation}
\label{eq10}
\frac{1}{n} \sum_{i=1}^n \lVert \nabla f_i(\tilde{w}) - \nabla f_i(w_*) \rVert ^2 \leq 2L(f(\tilde{w}) - f(w_*))
\end{equation}

\subsection{SVR-DQN in reducing AGE Variance}
We continue with Stochastic Variance Reduction for Deep Q-learning, recall the optimized gradient estimation procedure in Algorithm 1: 
\begin{align*}
g_{\mathrm{SVR-DQN}} &= w_m - \tilde{w}\\
&= w_{m-1} - \eta(\frac{1}{b}\sum_{i\in b}\nabla f_i(w_{m-1})\\
&-\frac{1}{b}\sum_{i\in b}\nabla f_i(\tilde{w})+\tilde{\mu}) - \tilde{w}\\
&= w_{m-2} - \cdots - \tilde{w}\\
&= w_0 - \sum_{i=0}^{m-1}\tau_i - \tilde{w}\\
&= - \sum_{i=0}^{m-1}\tau_i.
\end{align*} 
where $w_0 = \tilde{w}$ and $\tau_i$ denotes $\eta(\frac{1}{b}\sum_{j\in b}\nabla f_j(w_{i})-\frac{1}{b}\sum_{j\in b}\nabla f_j(\tilde{w})+\tilde{\mu})$. Since at each inner iteration mini-batch $b$ is uniform-sampled, for $i \neq j$: $\mathrm{Cov}(\tau_i,\tau_j) = 0$. Therefore, $\mathrm{Var}(g_{\mathrm{SVR-DQN}})=\sum_{i=0}^{m-1} \mathrm{Var}(\tau_i)$.

\begin{comment}
\begin{align*}
\mathrm{Var}(g_{\mathrm{SVR-DQN}}) &= \mathrm{Var}(-\sum_{i=0}^{m-1}\tau_i)\\
&=\sum_{i=0}^{m-1} \mathrm{Var}(\tau_i).
\end{align*}
\end{comment}

Furthermore, we have that $\mathbb{E}\lVert \tau_i \rVert^2 \leq \frac{8Lm\eta^2}{b}(f(\tilde{w}) - f(w_*))$(Appendix A), and therefore the following holds 
\begin{align*}
\mathrm{Var}(g_{\mathrm{SVR-DQN}}) \leq \frac{8Lm\eta^2}{b}(f(\tilde{w}) - f(w_*))\\
\mathrm{Var}(g_{\mathrm{Double-DQN}}) \leq \frac{2L}{B}(f(\tilde{w}) - f(w_*)),
\end{align*}
as shown in Appendix B and C, meaning that SVR-DQN is theoretically more efficient in AGE variance reduction than Double-DQN, and at least $\frac{b}{4m\eta^2B}$ times better than Double-DQN. Note that the the variance of $g_{\mathrm{SVR-DQN}}$ decreases as learning rate $\eta$ decreases theoretically. Whereas the $\eta$ shouldn't be too small or the sample efficiency can be too slow in practice. Therefore, a proper parameters setting should be carefully tuned, in our experiment setting  $B=512, b=32, m=32, \eta=0.01$ respectively.

\section{Experimental Results on Atari Games}
To demonstrate our method's effectiveness, we evaluate our proposed algorithm on a collection of $20$ games from Arcade Learning Environment~\cite{bellemare2013arcade}. This environment is considered as one of the most challenging datasets because of its high-dimensional state representation~\cite{van2016deep}. 
We utilize the similar neural network~\cite{mnih2015human} as the approximation of action value, taking raw images as input. The network architecture is a convolutional neural network with three convolutional layers and a fully-connected layer.
Following the paper ~\cite{mnih2015human}, we use the $\epsilon$-greedy scheme for exploration ,where $\epsilon$ is annealed linearly from 1.0 to 0.1 over the first million frames. All the experienced transitions are stored in a sliding replay memory, and the algorithm performs gradient descent on mini-batches of 512 transitions sampled uniformly from the reply memory. We set the learning frequency to 128, which means the training process repeats every 128 mini-batches. We also apply a frame-skipping strategy where the network takes the four frames as an input. All experiments are performed on an NVIDIA GTX Titan-X 12GB graphics card. In this paper, we utilize the tunned version of Double DQN algorithm~\cite{van2016deep}, as it somehow resolves the over-estimation issue in Q-learning.

\begin{figure*}
 \centering
  \begin{tabular}{@{}ccc@{}}
  
   \begin{minipage}{.3\textwidth}
    \includegraphics[width=\textwidth]{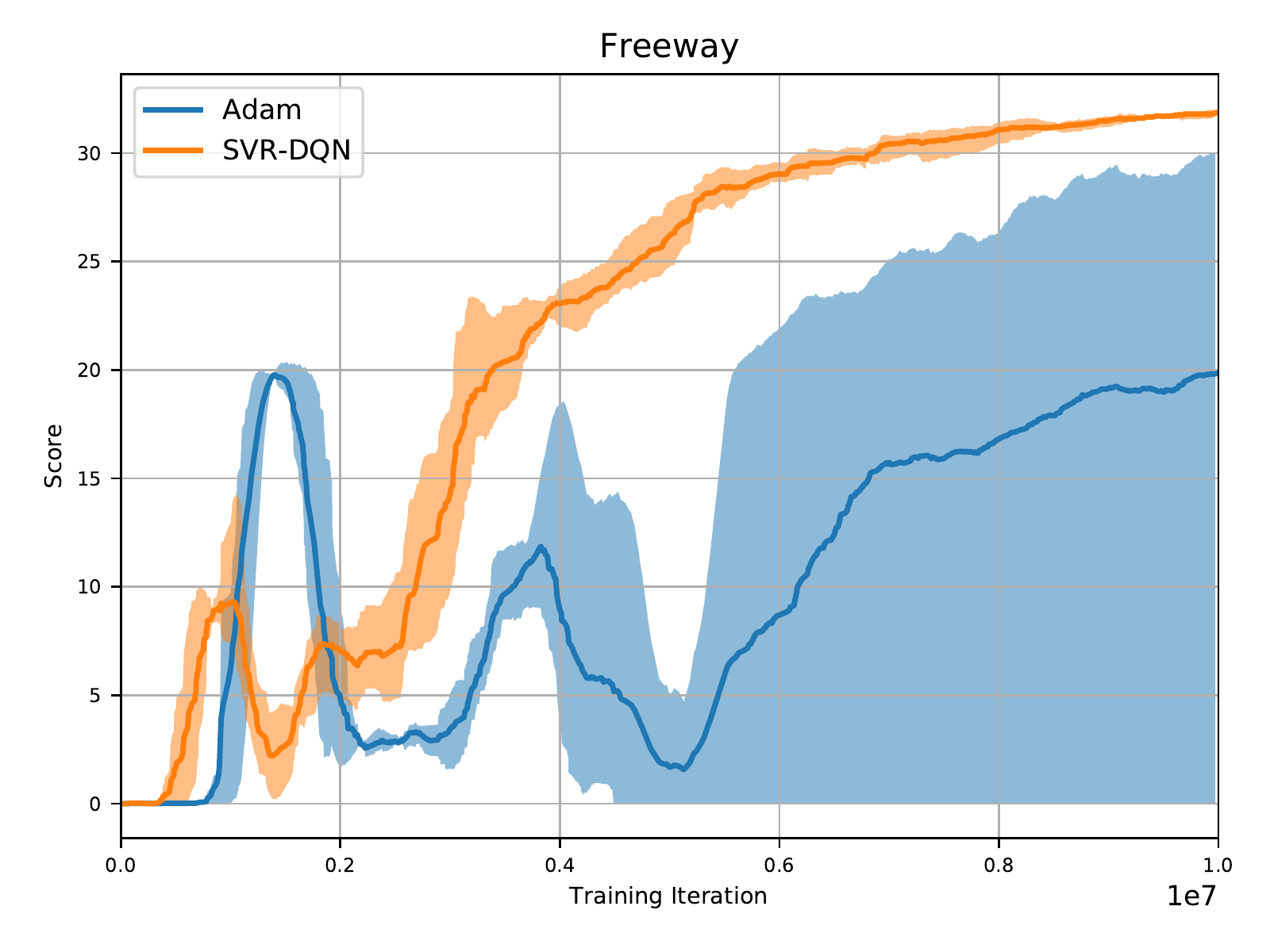}
   \end{minipage} &
    \begin{minipage}{.3\textwidth}
    \includegraphics[width=\textwidth]{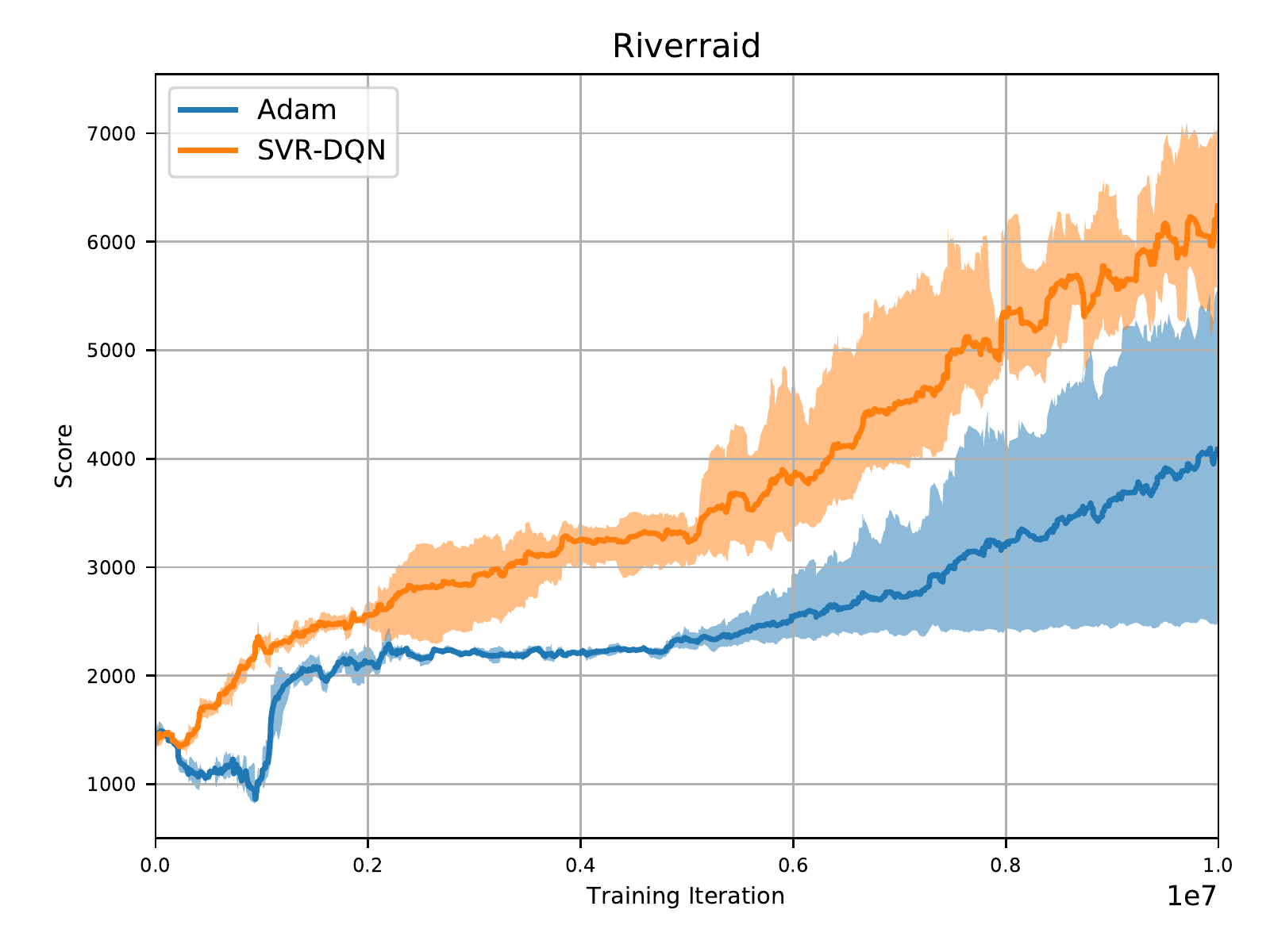}
   \end{minipage} &
      \begin{minipage}{.3\textwidth}
    \includegraphics[width=\textwidth]{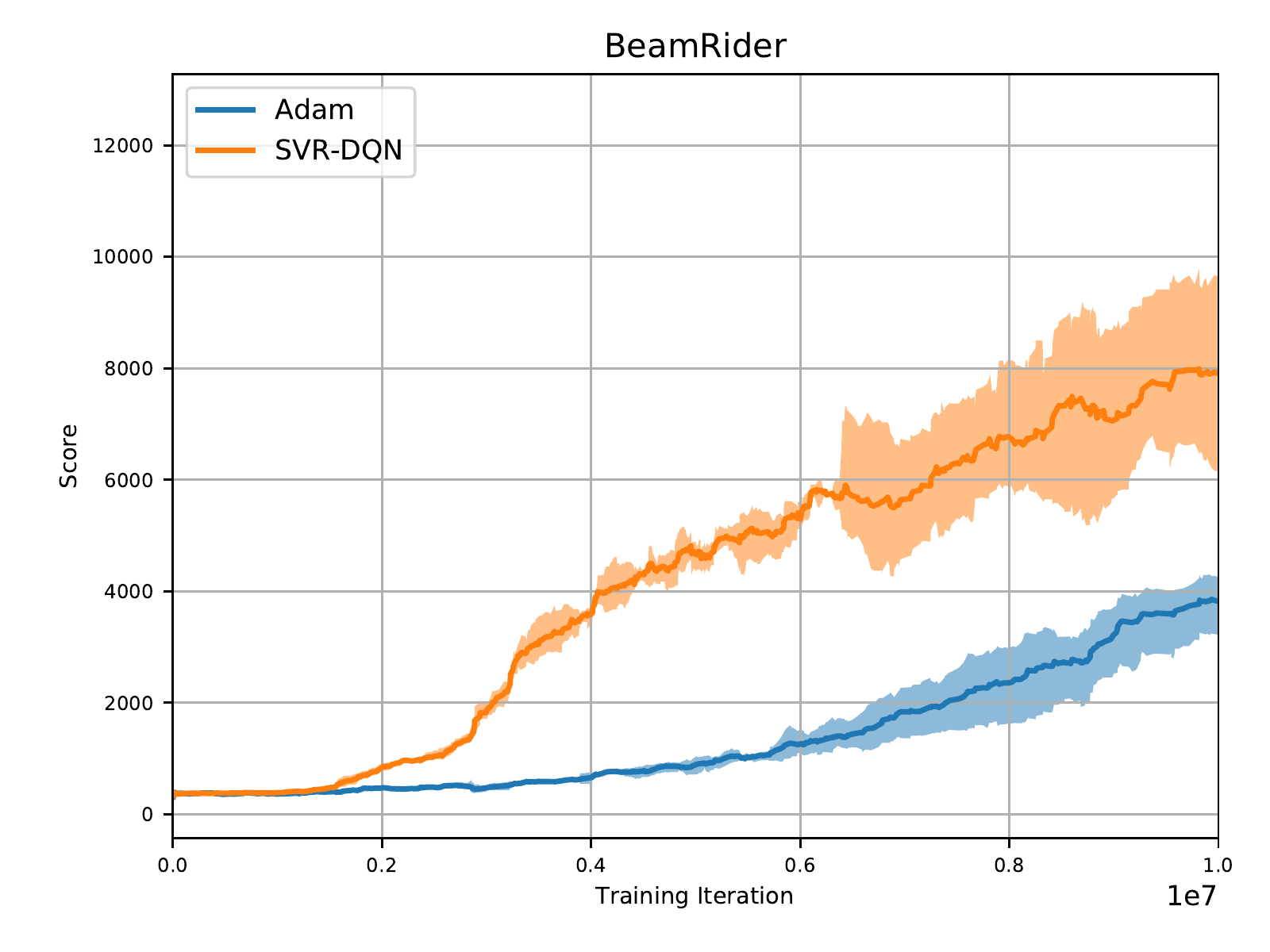}
   \end{minipage}\\
      \begin{minipage}{.3\textwidth}
    \includegraphics[width=\textwidth]{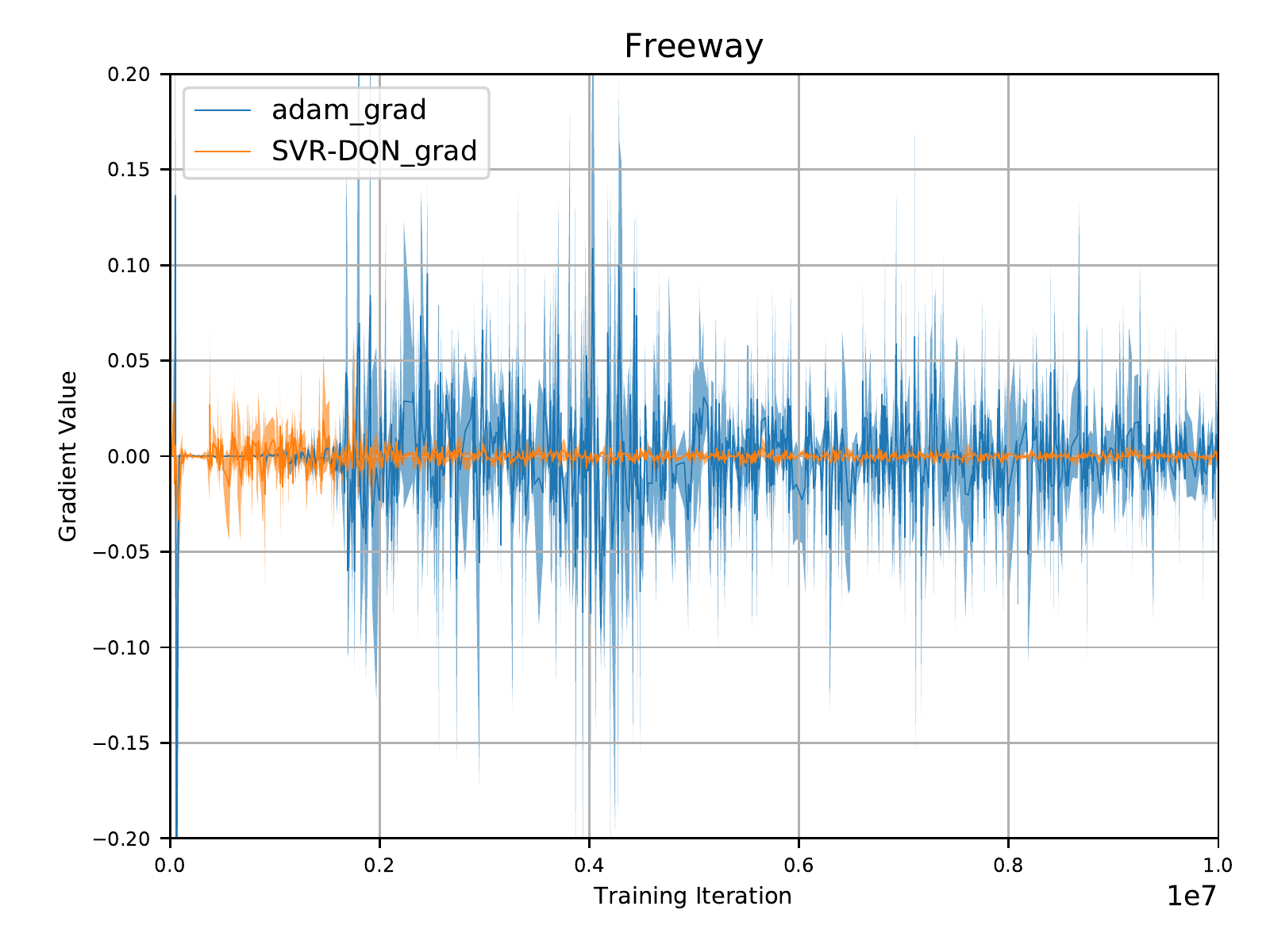}
   \end{minipage} &
    \begin{minipage}{.3\textwidth}
    \includegraphics[width=\textwidth]{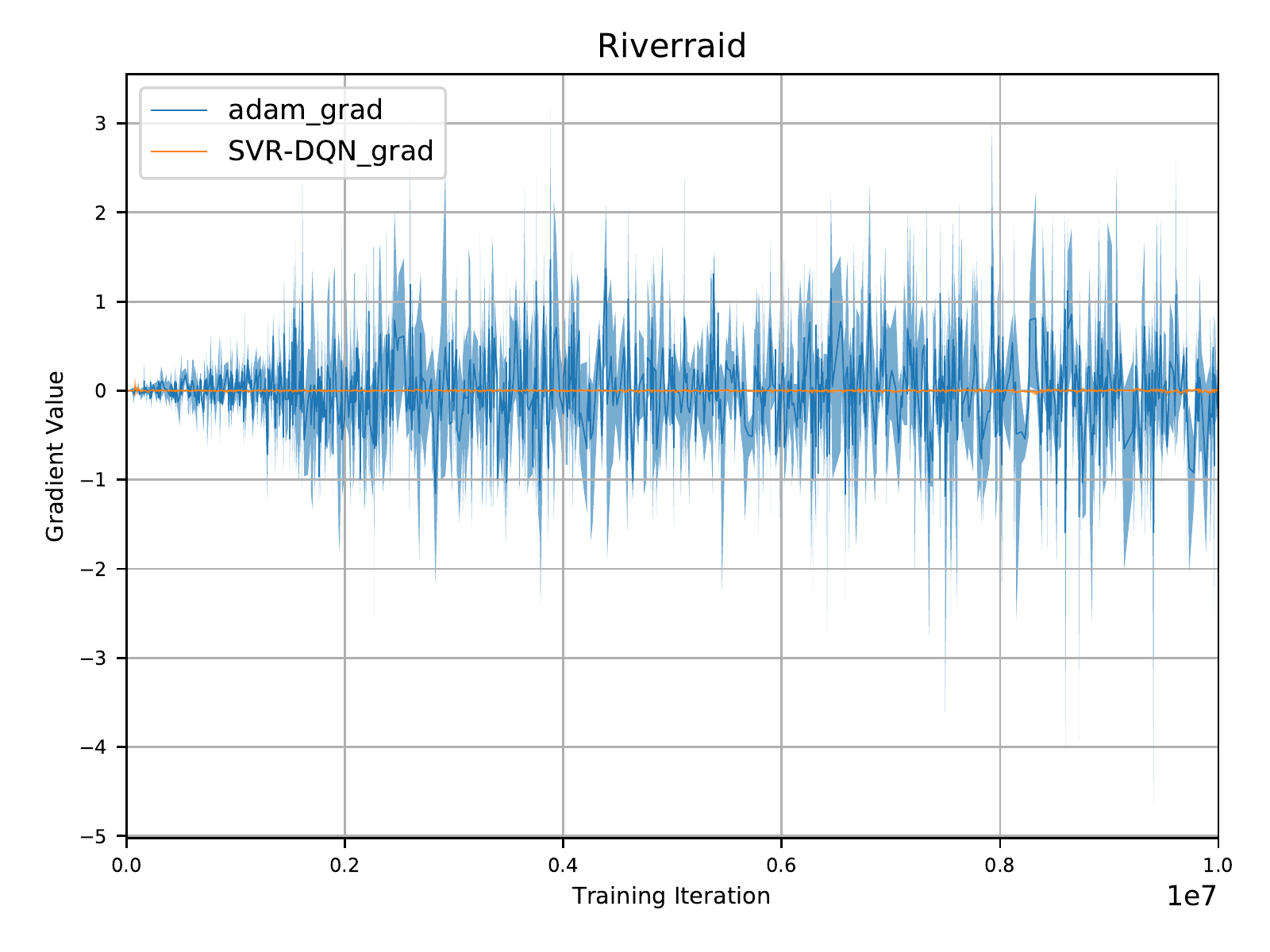}
   \end{minipage} &
      \begin{minipage}{.3\textwidth}
    \includegraphics[width=\textwidth]{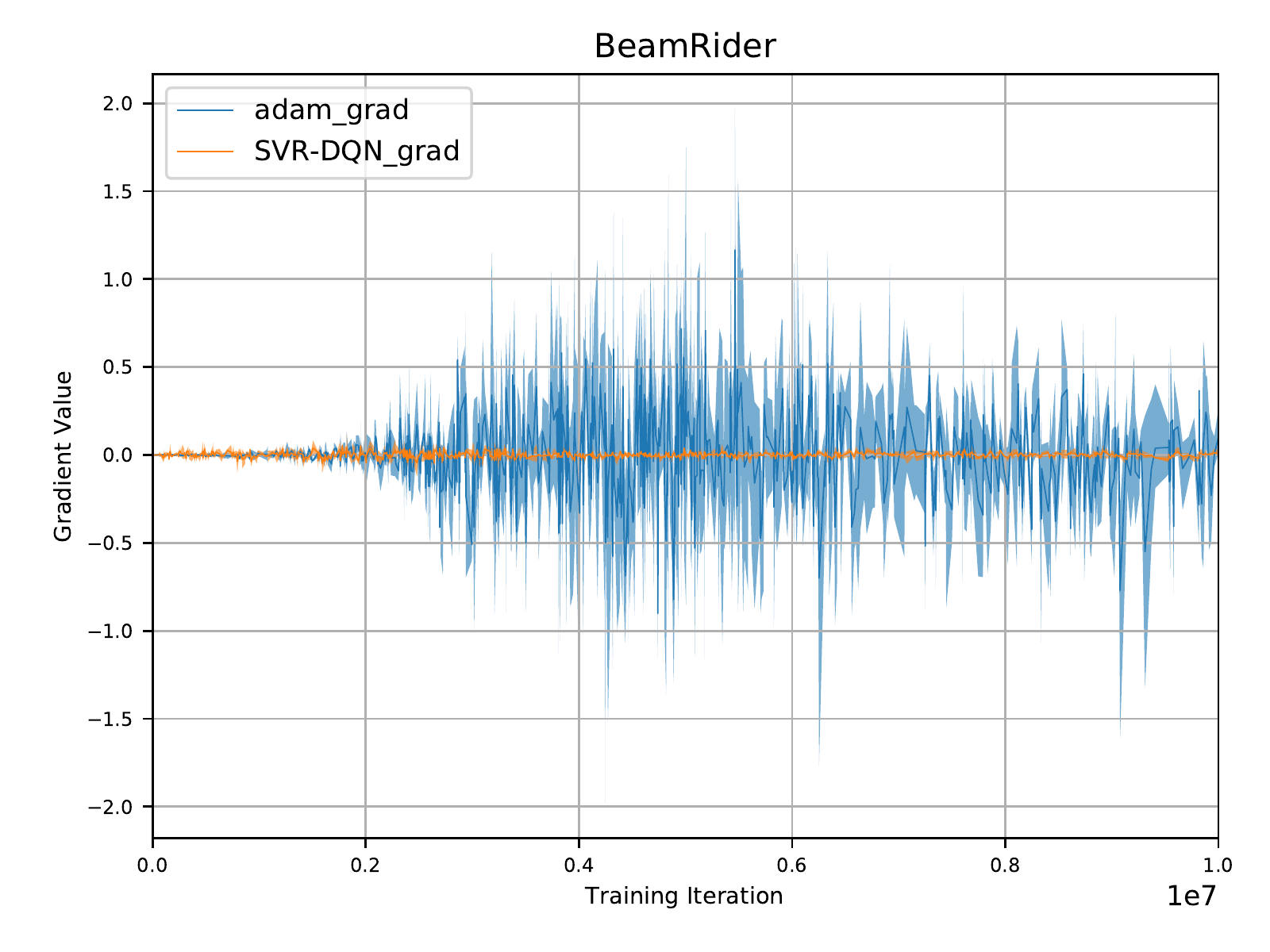}
   % \captionof*{figure}{somemore}
   \end{minipage}\\
  \end{tabular}
 \caption{The top row shows the learning curves (in raw score) for the Double DQN with Adam optimizer (blue), SVR-DQN optimizer (yellow), on 3 games of the Atari benchmark suite. The bold lines are averaged over 6 independent learning trials (6 different seeds).The performance test using $\epsilon$-greedy policy with 10 million iterations. The shaded area presents one standard deviation. The bottom row shows that when applied SVR-DQN, the variance of averaged gradient estimation is largely reduced, performance improves, and less variability is observed. 30 no-op evaluation is used and moving average over 4 points is applied. Here x-axis denotes the number of training frames while y-axis denotes the evaluation score in the game. }
 \label{detailed}
\end{figure*}

\subsection{Evaluation}
Our proposed algorithm can obtain more accurate gradient estimation through the same batch of training samples compared to baseline, and more accurate gradient evaluation should result in more aggressive learning curves in the initial training stage. Though the previous work~\cite{mnih2015human} trained their agent using 200 million (200M) frames or 50M training iteration for each game, we choose to train our agent within only 40M frames or 10M training iterations, due to time constraints. Note that regarding evaluating the performance of SVR-DQN, our main concerns focus on the performance in initial stage. Instead of using Double DQN baseline results for those 20 games published from previous work, to obtain fair comparison, we replicate the baseline results using the same hyper-parameter setting, code base, and random seed initialization as SVR-DQN for $10$M training frames. The only difference is the optimizer we utilized to minimize the bellman error, where our gradient estimator could lead to a smaller variance. Our experiments could be finished within two days. 

\begin{figure}
  \includegraphics[scale=0.75]{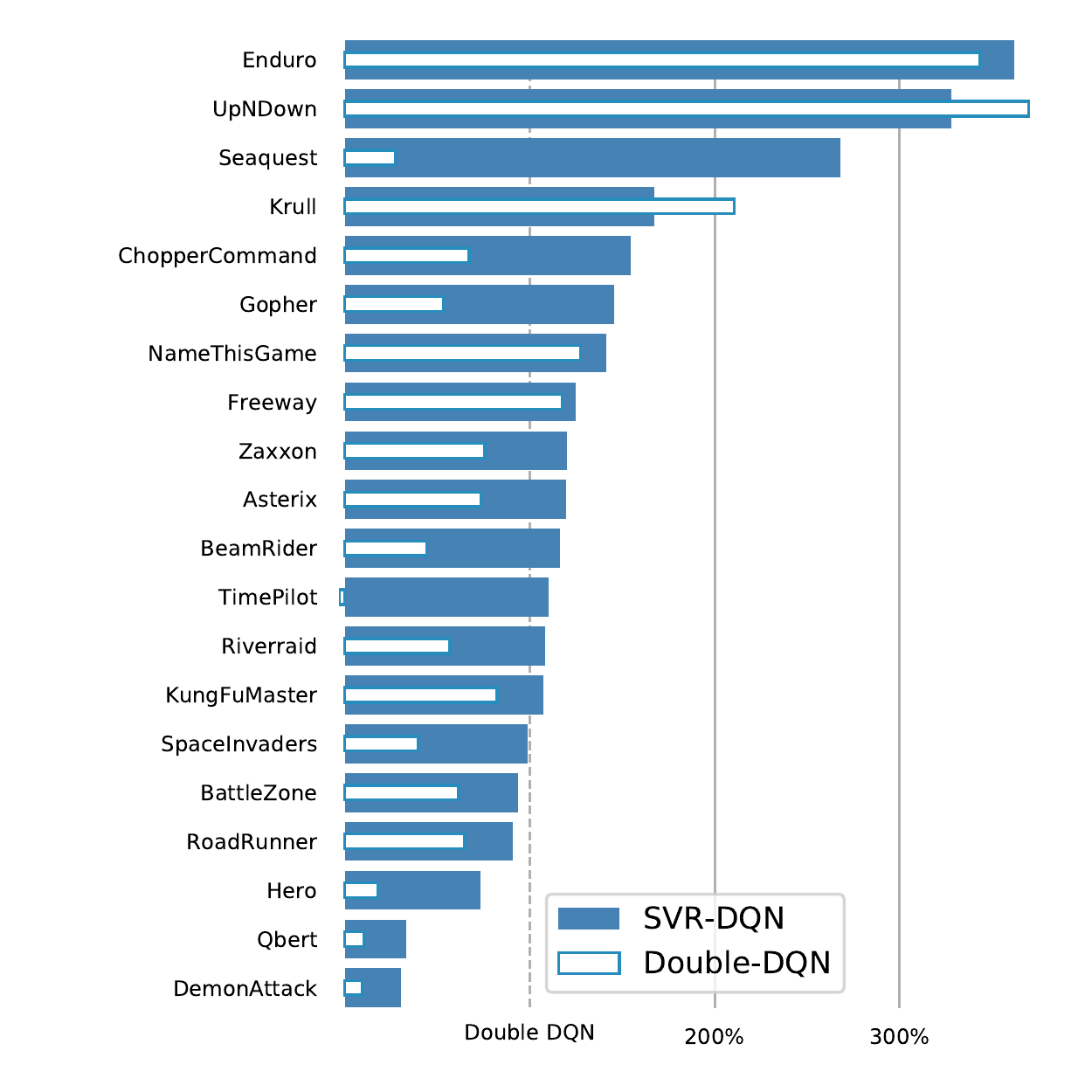}
  \caption{Normalized score on 20 Atari games, tested for $100$ episodes per game. The blue bars denotes our SVR-DQN while the white bars denote the Adam optimizer, which is a baseline.}
  \label{normalized}
\end{figure}

Our evaluation procedure follows the description by~\cite{mnih2015human}, we apply `30 no-op evaluation' to provide different starting points for the agent.  Our agent is evaluated after a maximum of 5 minute gameplay, which contains $18,000$ frames, with the usage of $\epsilon$-greedy policy where $\epsilon=0.05$. The rewards are the average from 100 episodes. For each game, our agent is evaluated at the end of every epoch ($160$ epochs in total). To compare the performance of our algorithm to the Double DQN baseline across games, we apply the normalization algorithm proposed by~\cite{van2016deep} to obtain the normalized improvement score in percent as follows:
 \begin{equation}
\text{score}_{\text{normalized}} = \frac{\text{score}_{\text{agent}}-\text{score}_{\text{random}}}{|\text{score}_{\text{Double DQN}}-\text{score}_{\text{random}}|}
\label{norm}
 \end{equation}
 \noindent The detailed results could be found in Fig. \ref{normalized} and Table \ref{summarize}.
 
\begin{table}
 \centering
   \begin{tabular}{|c|cc|}
   \hline
   & Mean & Median\\
   \hline
   SVR-DQN & \bf{139.75\%} & \bf{118.02\%} \\
   \hline
   Double DQN & 92.48\% & 63.13\% \\
   \hline
   \end{tabular}
 \caption{Mean and median normalized scores.}
 \label{summarize}
\end{table} 

In summary, we adopt the `Double DQN' and `random' score reported by~\cite{mnih2015human}, the results are demonstrated in Fig. \ref{normalized}. We observe a better performance on 19 out of 20 games, which demonstrates the effectiveness of our proposed algorithm. We also give the summary statistics in terms of mean and median score in Table \ref{summarize}. Compared to Adam, the median performance across 20 games increases from $63.13$\% to $118.02$\% and the mean performance increases from $92.48$\% to $139.75$\%. Noteworthy examples include Seaquest (from $27.42$\% to $267.94$\%), Gopher (from $53.22$\% to $145.42$\%).

\begin{figure}
 \centering
  \includegraphics[scale=0.4]{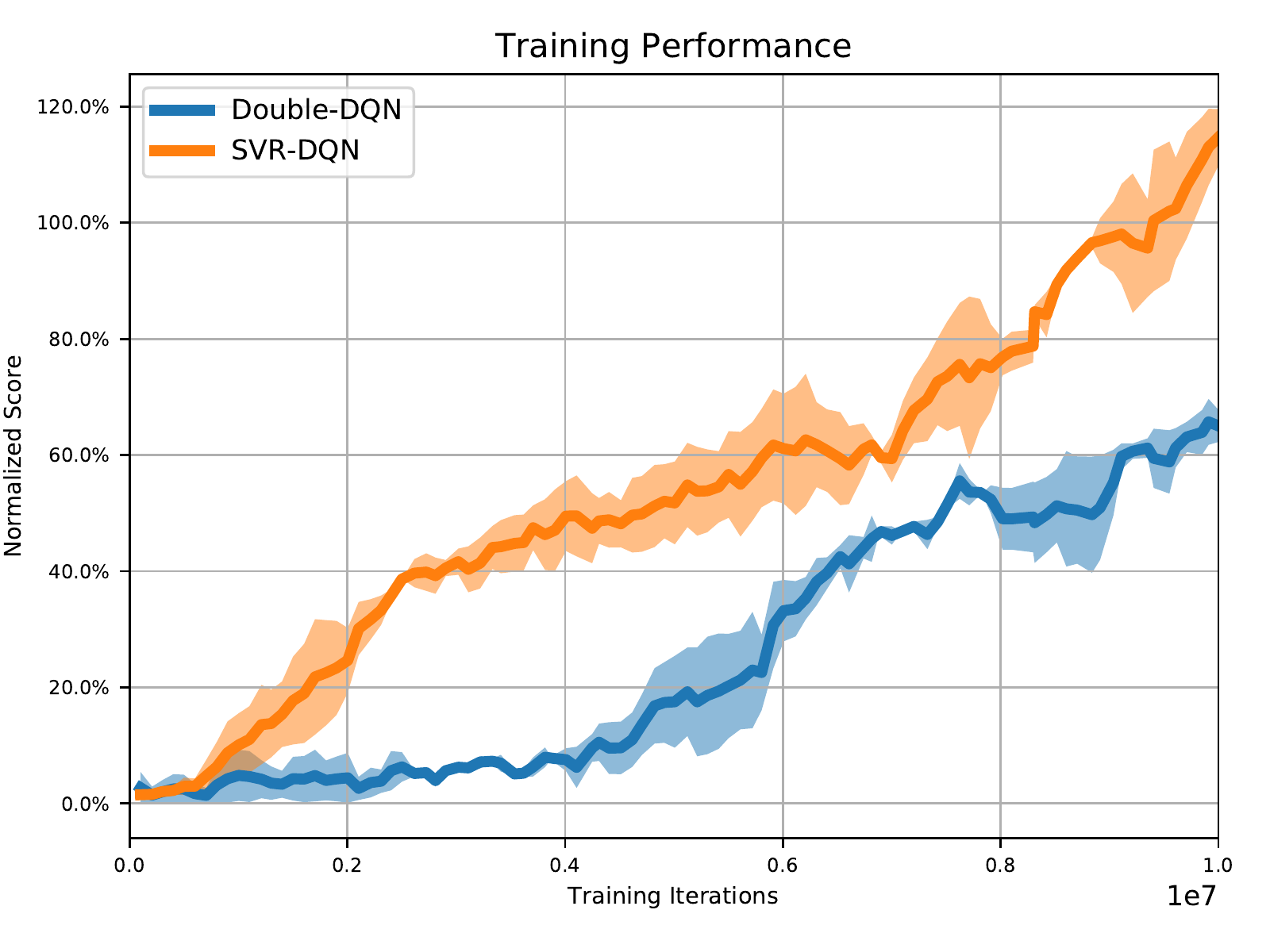}
  \caption{Summary plots of sample efficiency. Median over 20 games of the normalized score achieved so far. The normalized score is calculated in Equation \ref{norm}.}
  \label{speed}
\end{figure}

We also conduct a comparison of the sample efficiency and results could be found in Fig.\ref{speed}.  We observe that SVR-DQN boosts the performance on almost all games, and the sample efficiency of SVR-DQN is nearly twice as fast as original Double-DQN with Adam optimizer. 

Also, the performances of $3$ representative games are reported in Fig.\ref{detailed}. The three games include `BeamRider', `Freeway', `Riverraid'. As can be seen in Fig.\ref{detailed} that out proposed SVR-DQN method results in significant lower average gradient estimates, and the variance of gradient is largely reduced. We also observe that our method outperforms the baselines with a significant margin on the majority of the games, and SVR-DQN leads to less variability between the runs of independent learning trials. For the game of Freeway, we see that the divergence of Double-DQN can be prevented by SVR-DQN. On the other hand, the performance of Double-DQN with Adam optimizer has a sudden deterioration at 4M iteration where the gradient variance suddenly increases. 

It is noteworthy that SVR-DQN usually leads to aggressive performance improvement at the initial training stage. We believe that our method can be combined with other techniques developed for DQN, such as prioritized experience replay~\cite{schaul2015prioritized}, dueling networks~\cite{wang2015dueling} to further improve the effectiveness.

%SVR-DQN significantly reduces the delay before the performance gets off the ground in games that otherwise suffer from such delay. 
\subsection{Understanding SVR-DQN}
We take the investigation of our SVR-DQN algorithm in terms of the performance improvement brought by variance control technique and computation acceleration impacts. Notice all the experimental settings are pre-described in section 5 and are kept constant through all the experiments. 

From Fig.\ref{detailed}, it is clear that the major weapon helping SVR-DQN outperform significantly than Double-DQN is the usage of stochastic variance reduction strategy. Here the Double-DQN denotes we solely use Adam optimization with the vanilla gradient estimation, this choice allows for improvement in computation efficiency but causes a larger estimation variance due to mini-batch estimation noise, long horizon noise and unknown dynamics, etc. The reduction of the AGE variance is crucial for achieving faster convergence rate. As you can see for Freeway(Fig.\ref{detailed}), Double-DQN with vanilla gradient estimation gets stuck at bad local optimal and difficult for the performance to get off the ground due to high variance inaccurate estimation. It can be beneficial for exploration of the parameter space and better performance around current policy with small gradient noise being controlled. However, if the gradient variance is very wild, the performance can be damaged greatly as illustrated in all the three games from Fig.\ref{detailed}

Another major weapon is the subsampling strategy to accelerate the computation speed as illustrated throughout our experiments. Proved by ~\cite{harikandeh2015stopwasting}, if we assume the sample variance of the gradients norms is bounded for each iteration,  the convergence rate is the constant with the usage of full batch, when the sub-batch size $|B|$ is properly selected. Using subsampling strategy as a building block, we further propagate it into SVRG inner loop using mini-batch sample instead of individual training sample which has large variance among each other. Note that the sample complexity is what we concern more in reinforcement learning experiments, the computation cost of stochastic variance reduction step is negligible compared to simulation time, which is also confirmed throughout our experiments that both SVR-DQN and Double for 10 million iteration can be finished within 2 days across all the tested games, there is no significant training time difference between the two methods. 

\section{Related Work}
In recent years, numerous %of
techniques have been proposed to improve the convergence and stability of deep reinforcement learning and optimization method plays a critical role. The well-known \textbf{REINFORCE}~\cite{williams1992simple} uses SGD method. To accelerate the convergence rate and solve the challenges aforementioned, some important improvements are explored. AdaGrad~\cite{duchi2011adaptive} adapts learning rate with respect to the frequency of parameters, and is well-suited for dealing with sparse data. RMSprop~\cite{tieleman2012lecture} improves AdaGrad by resolving its radically diminishing learning rates. Adaptive Moment Estimate (Adam)~\cite{kingma2014adam} combines the advantage of both AdaGrad and RMSprop while keeping momentum technique, %Adam
empirically outperforming other adaptive learning-method algorithms. Under the mechanics of variance control, representative methods such as SAG~\cite{roux2012stochastic} and SDCA~\cite{shalev2013stochastic} are proposed,.%Stochastic Variance Reduction~\cite{johnson2013accelerating} significantly outperforms SAG and SDCA in terms of convergence rate with dramatic variance reduction and doesn't require large gradient storage.
In terms of convergence rate with dramatic variance reduction and not requiring large gradient storage, Stochastic Variance Reduction~\cite{johnson2013accelerating} significantly outperforms SAG and SDCA. Recently, second-order statistics optimization algorithms are adopted~\cite{battiti1992first}~\cite{wang2017improved}. However, second-order methods are infeasible in practice for high-dimensional training, such as neural network.

\subsection{Stochastic Gradient-based Optimization}
By far, stochastic gradient descent is 
%the most
a common method for neural networks optimization~\cite{kingma2014adam}. Many optimization problems can be summarized as  finding the minima or maxima of scalar objective function $J(\theta)$. Gradient descent updates the parameters in the opposite direction of the gradient of $J(\theta)$ until reaching a minimum. However, objective functions are often stochastic %where
as they are composed of different subfunctions~\cite{kingma2014adam}. In such cases, stochastic gradient descent (SGD) improves gradient descent by computing gradients with a single or a few training examples and takes gradient steps through individual subfunctions~\cite{bottou2010large}.  Although SGD exhibited its efficiency in many machine learning 
%success
successful stories, there are still some key challenges wait to be solved, including choosing a proper learning rate schedule and %how to 
avoiding to get trapped in numerous suboptimal local minima in non-convex neural networks training~\cite{choromanska2015loss}. Therefore, efficient stochastic optimization techniques are required.

\subsection{Variance Reduction in Deep Q Learning}
In addition to optimization algorithms, numbers of techniques are also proposed to reduce varieties of variance in deep Q learning. The well-known variance in DQN is the Q learning overestimation error, which is first investigated by \cite{baird1993advantage}, who has showed that since action values contain random errors distributed in the interval $[-\epsilon, \epsilon]$. Since the DQN target is obtained using max operator, the expected overestimation errors are bound by $\gamma \in \frac{n-1}{n+1}$, where $n$ is the applicable action numbers given current state $s$. the intuition nature of overestimation error is that it can cause asymptotically sub-optimal policies, as shown by~\cite{baird1993advantage} and later by~\cite{van2016deep} that noisy in Arcade Learning Environment can lead to overestimation. The Double DQN~\cite{van2016deep} is a possible way to tackle overestimation error which replaces the positive bias with a negative one, where two Q-network are applied for Q action selection and Q function value calculation respectively. 

Another variance in DQN is the Target Approximation Error (TAE), which is investigated by~\cite{anschel2016averaged} where TAE is result of sub-optimality of Q function parameter $\theta$ due to inexact minimization and limited representation power of DQN. A efficient method to reduce TAE variance is Average DQN~\cite{anschel2016averaged}, the key idea is to use the $K$ previously calculated Q-values to estimate the current action-value estimate. Another recent explored variance is caused by reward signals noise, which is investigated by~\cite{romoff2018reward}. In order to reduce reward signal variance, a direct reward estimator $\hat{R}(s_t)$ is proposed to update the discounted value function instead of sampled reward.
Our stochastic variance reduction for deep Q learning method differs from all of the aforementioned approaches. The key idea of our method is to reduce the variance caused by approximate gradient estimation, and thus greatly improve the efficiency and performance.

\section{Conclusion}
In this paper we proposed an innovative optimization algorithm for Q-learning which reduces the variance in gradient estimation, our proposed optimization algorithm achieves significantly faster convergence than the Adam optimizer. Our method significantly improves the performance of Double DQN on the Atari 2600 domain. In the future, we plan to investigate the impact advanced constrained optimization and explore the potential synergy with other techniques.

\begin{comment}
\section*{Acknowledgments}
\end{comment}
%% The file named.bst is a bibliography style file for BibTeX 0.99c

\clearpage
\bibliographystyle{aaai}
\bibliography{SVR-DQN}

\begin{comment}
\clearpage
\appendix
\section{Additional Results}\label{stylefiles}
\begin{figure*}
 \centering
  \begin{tabular}{@{}cccc@{}}
    \includegraphics[width=.23\textwidth]{Asterix.pdf} &
    \includegraphics[width=.23\textwidth]{BattleZone.pdf} &
    \includegraphics[width=.23\textwidth]{BeamRider.pdf} &
    \includegraphics[width=.23\textwidth]{ChopperCommand.pdf}\\
    \includegraphics[width=.23\textwidth]{DemonAttack.pdf} &
    \includegraphics[width=.23\textwidth]{Enduro.pdf} &
    \includegraphics[width=.23\textwidth]{Freeway.pdf} &
    \includegraphics[width=.23\textwidth]{Gopher.pdf}\\
    \includegraphics[width=.23\textwidth]{Hero.pdf} &
    \includegraphics[width=.23\textwidth]{Krull.pdf} &
    \includegraphics[width=.23\textwidth]{KungFuMaster.pdf} &
    \includegraphics[width=.23\textwidth]{NameThisGame.pdf}\\
    \includegraphics[width=.23\textwidth]{Qbert.pdf} &
    \includegraphics[width=.23\textwidth]{Riverraid.pdf} &
    \includegraphics[width=.23\textwidth]{RoadRunner.pdf} &
    \includegraphics[width=.23\textwidth]{Seaquest.pdf}\\
    \includegraphics[width=.23\textwidth]{SpaceInvaders.pdf} &
    \includegraphics[width=.23\textwidth]{TimePilot.pdf} &
    \includegraphics[width=.23\textwidth]{UpNDown.pdf} &
    \includegraphics[width=.23\textwidth]{Zaxxon.pdf}\\
  \end{tabular}
 \caption{Detaile learning curves (in raw score) for Double DQN with Adam optimizer (blue), SVR-DQN optimizer (yellow), for all 20 games of the Atari benchmark suite.}
 \label{hiya}
\end{figure*}
\end{comment}

\end{document}